\documentclass[final]{cvpr}

\usepackage{times}
\usepackage{epsfig}
\usepackage{graphicx}
\usepackage{amsmath}
\usepackage{amssymb}
\usepackage[ruled,vlined,linesnumbered]{algorithm2e}
\usepackage{multirow}
\usepackage{arydshln}
\usepackage{booktabs}
\DeclareMathOperator*{\argmin}{arg\,min}
\pdfoutput=1

\usepackage[pagebackref=true,breaklinks=true,colorlinks,bookmarks=false]{hyperref}

\begin{document}

\title{KCP: Kernel Cluster Pruning for Dense Labeling Neural Networks}

\author{Po-Hsiang Yu\\
	National Taiwan University\\
	{\tt\small r08943024@ntu.edu.tw}
	\and
	Sih-Sian Wu\\
	National Taiwan University\\
	{\tt\small benwu@video.ee.ntu.edu.tw}
	\and
	Liang-Gee Chen\\
	National Taiwan University\\
	{\tt\small lgchen@ntu.edu.tw}
}

\maketitle

\begin{abstract}
   Pruning has become a promising technique used to compress and accelerate neural networks. Existing methods are mainly evaluated on spare labeling applications. However, dense labeling applications are those closer to real world problems that require real-time processing on resource-constrained mobile devices. Pruning for dense labeling applications is still a largely unexplored field. The prevailing filter channel pruning method removes the entire filter channel. 
   Accordingly, the interaction between each kernel in one filter channel is ignored.
   
   In this study, we proposed kernel cluster pruning (KCP) to prune dense labeling networks. We developed a clustering technique to identify the least representational kernels in each layer. By iteratively removing those kernels, the parameter that can better represent the entire network is preserved; thus, we achieve better accuracy with a decent model size and computation reduction. When evaluated on stereo matching and semantic segmentation neural networks, our method can reduce more than 70\% of FLOPs with less than 1\% of accuracy drop. Moreover, for ResNet-50 on ILSVRC-2012, our KCP can reduce more than 50\% of FLOPs reduction with 0.13\% Top-1 accuracy gain. Therefore, KCP achieves state-of-the-art pruning results.
\end{abstract}

\section{Introduction}
\begin{figure}[t!]
	\centering
	\includegraphics[scale=0.4]{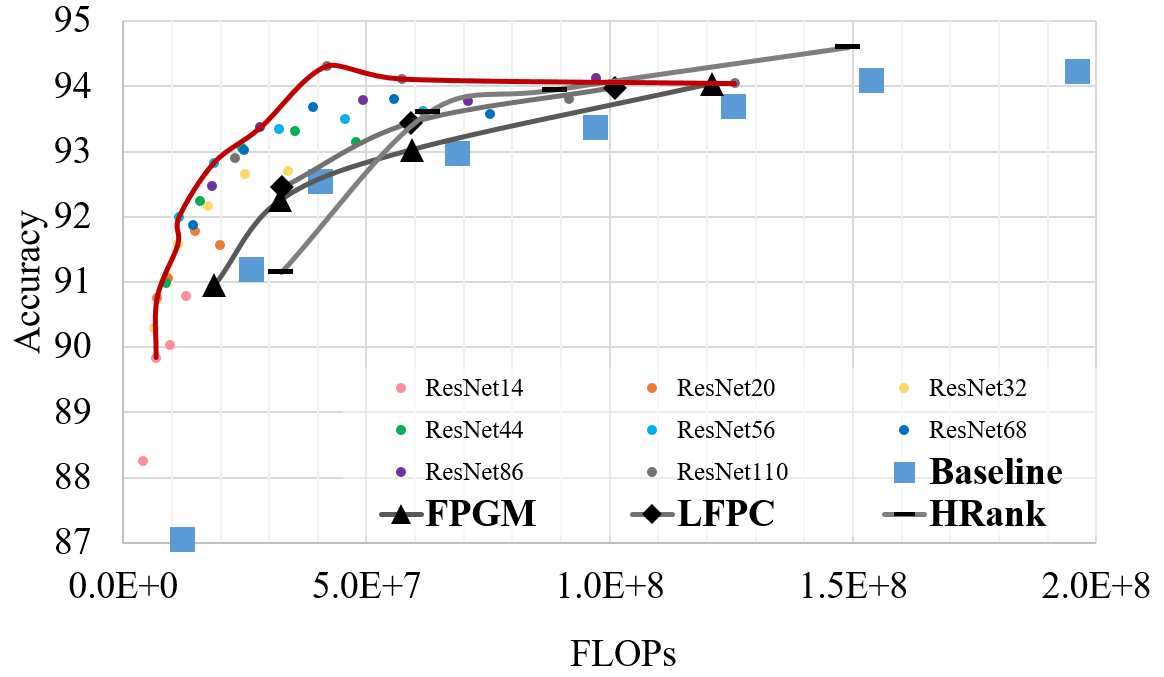}
	\caption{KCP benchmarked on CIFAR-10. Points that is located near upper left are preferred because they achieved better accuracy-FLOPs trade-off. Baseline points are the original ResNet. Lines are the envelopes of points from different depth ResNet. Our method achieves better efficiency than previous state-of-the-arts, namely, FPGM~\cite{he2019GM}, LFPC~\cite{he2020LFPC}, and HRank~\cite{lin2020hrank}. Model pruned by KCP are closer to the upper left corner, and our envelope is decently above that of all other works.}
	\label{fig:cifar_intro}
\end{figure}
\begin{figure}[t!]
	\centering
	\includegraphics[scale=0.22]{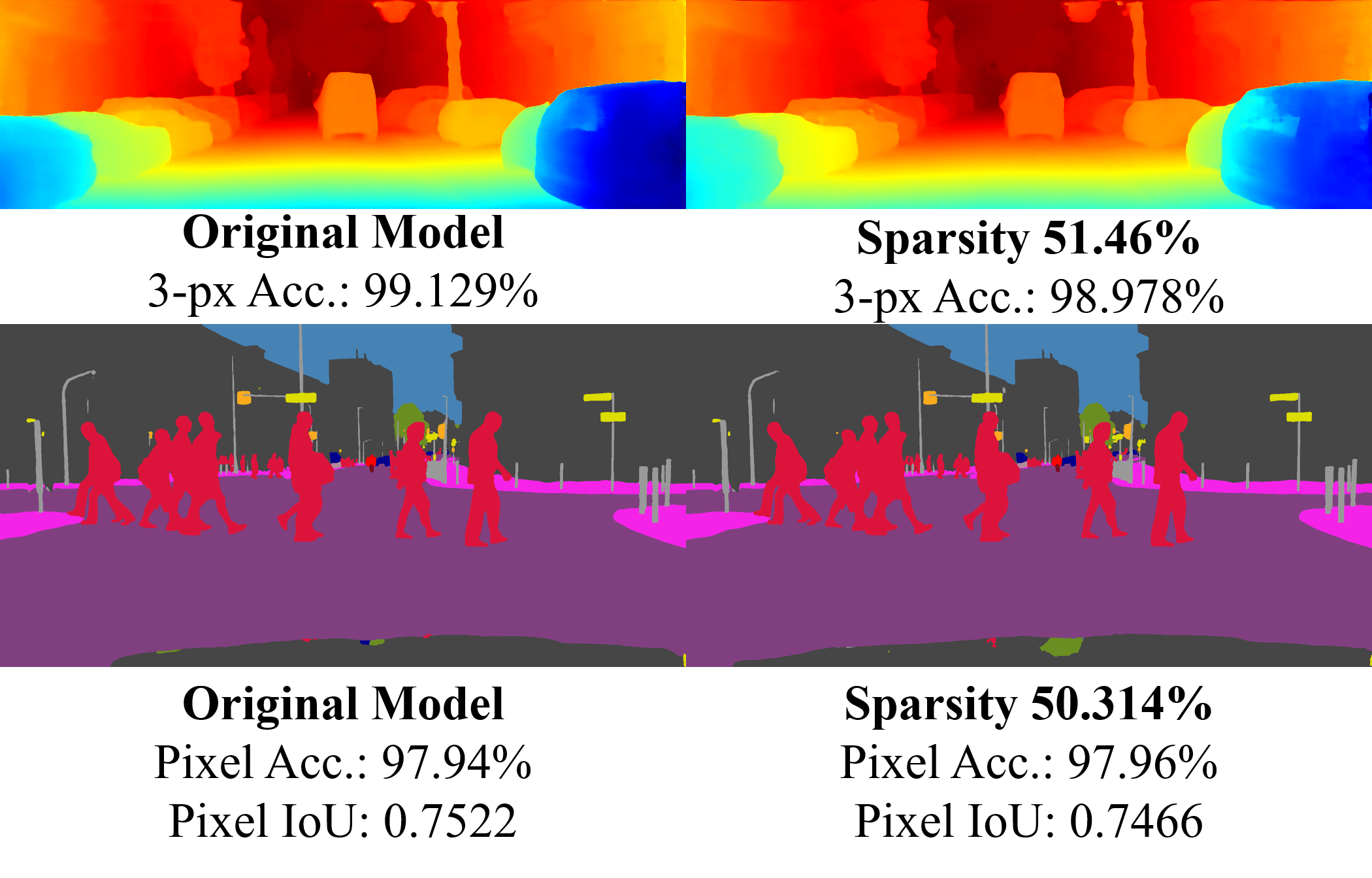}
	\caption{Results of KCP on pruning dense labeling networks. The upper row show the results of depth estimation; the predicted depth are samples from KITTI2015~\cite{KITTI2015C, KITTI2015J} generated by the original PSMNet~\cite{PSMNet} and the KCP processed 50\% sparsity model. The lower row shows the result of semantic segmentation; the predicted class label are samples from Cityscapes~\cite{Cordts2016Cityscapes} generated by original HRNet~\cite{HRNet} and KCP processed 50\% sparsity model. KCP maintains the model performance both quantitatively and qualitatively.}
	\label{fig:intro_vis}
\end{figure}

Deep learning based algorithms have been evolving to achieve significant results in solving complex and various computer vision applications, such as image classification~\cite{resnet, krizhevsky2012imagenet, simonyan2014very}, stereo matching~\cite{vzbontar2016stereo, PSMNet}, and semantic segmentation~\cite{long2015fully, chen2017deeplab, HRNet}. However, real world applications often require the deployment of these algorithms on resource-limited mobile/edge devices. Owing to over-parameterization and high computational cost, it is difficult to run those cumbersome models on small devices~\cite{yu2020joint}, thereby hindering the fruits of those state-of-the-art (SOTA) intelligent algorithms from benefiting our lives. Neural network pruning is a technique to strike the balance between accuracy (performance) and cost (efficiency). Filter level pruning has been proven to be an effective~\cite{li2016pruningfilter, luo2017thinet, he2019GM, lin2020hrank, guo2020dmcp, he2020LFPC} and favorable method because it could result in a more structural model.

Despite the numerous filter pruning algorithms that have been proposed, previous works are mostly evaluated on a sparse labeling classification task. In a practical scenario, dense labeling problems often require real-time processing in real world problems. Hence, the compression of a dense labeling neural network is crucial and demanding, yet this field is still quite unexplored. Moreover, filter channel pruning removes the entire output channel, which can sometimes lead to sub-optimal results in terms of accuracy preserving. A single filter channel often contains several kernels; thus, the information density is still high. If one removes the entire channel, the interactions between each kernels are neglected. In such a situation, each kernel in one filter may contribute differently to the final prediction. Pruning at the level of the channel is equivalent to aggregating the influence of an individual kernel, which dilutes the uniqueness of each kernel. 

We propose kernel cluster pruning (KCP) to address the issues and fill the gap in dense labeling pruning. The main concept of KCP is to identify the least representational kernels in each layer and subsequently remove them. The smallest target to be pruned in KCP is the kernel instead of the filter. First, the cluster center of kernels in each layer are calculated. Thereafter, the kernels that are closest to the cluster center are considered to carry less information and are removed. Because every pixel in the prediction is assigned a label, dense labeling networks are more vulnerable to change in parameter than sparse labeling networks. Kernel pruning enables us to compress dense labeling networks more delicately while still maintaining the network structure after pruning. Experiments showed that we not only successfully compressed the dense labeling network but also achieved SOTA results on CIFAR-10~\cite{cifar10} and ImageNet (ILSVRC-2012)~\cite{imagenet}. 

Our main contributions are summarized as follows:
\begin{itemize}
	\item A novel kernel pruning method, KCP, is proposed. This method can identify the kernels with least representativeness in each layer then prune those kernels without breaking the original network structure.
	\item Thorough evaluations of dense labeling neural network pruning are presented. The results show that KCP can effectively prune those networks and retain the accuracy. To the best of our knowledge, this is the first work that has investigated structured filter pruning on dense labeling applications.
	\item The experiment on benchmark datasets demonstrates the effectiveness of KCP. KCP achieves new SOTA results for ResNet on CIFAR-10 (c.f. Fig~\ref{fig:cifar_intro}, Table~\ref{table:cifar}) and ImageNet (c.f. Table~\ref{table:imagenet}).
\end{itemize}

\section{Related Works}
The prior art on network pruning can be approximately divided into several approaches, depending on the pruning objectives and problem formulation. 
The early pruning methods are individual \emph{weight pruning}. The idea can first be observed in optimal brain damage~\cite{lecun} and optimal brain surgeon~\cite{hassibi1993}. More recently, Han \etal.~\cite{han2015learn} proposed a series of works removing connections with weights below a certain threshold and later incorporated the idea into the deep compression pipeline~\cite{han2015deep}. In ~\cite{liu2018rethinking} (Liu \etal), the results indicated that the value of automatic structured pruning algorithms sometimes lie in identifying efficient structures and performing an implicit architecture search, rather than selecting “important” weights. Different from rethinking~\cite{liu2018rethinking} ,the "lottery ticket hypothesis", proposed by Frankle \etal~\cite{frankle2018lottery} is another interesting work about the reason for pruning. They found that a standard pruning technique naturally uncovers sub-networks whose initializations made them capable of being trained effectively.  

Meanwhile, \emph{structured} pruning methods prune the entire channel or filter away. Channel pruning has become one of the most popular techniques for pruning owing to its fine-grained nature and ease of fitting into the conventional deep learning framework. 
Some heuristic approaches use a magnitude-based criterion to determine whether the filter is important. Li \etal~\cite{li2016pruningfilter} calculated the $L1$-norm of filters as an important score and directly pruned the corresponding unimportant filter. Luo \etal~\cite{luo2017thinet} selected unimportant filters to prune based on the statistics information computed from its subsequent layer. He \etal~\cite{he2018SoftFP} proposed to allow pruned-away filters to recover during fine-tuning while still using $L1$-norm criterion to select the filter. Molchanov \etal ~\cite{molchanov2019importance} further proposed a method that measures squared loss change induced by removing a specific neuron, which can be applied to any layer. He \etal~\cite{he2019GM} calculated the geometric median of the filters in the layer and removed the filter closest to it. Lin \etal~\cite{lin2020hrank} discovered that the average rank of multiple feature maps generated by a single filter is always the same, regardless of the number of image batches convolution neural networks (CNNs) receive. They explored the high rank of feature maps and removed the low-rank feature maps to prune a network. He \etal.~\cite{he2020LFPC} developed a differentiable pruning criteria sampler to select different pruning criteria for different layers. Group sparsity imposed on sparsity learning was also extensively used to learn the sparse structure while training a network. Some used $L1$-regularization (Liu \etal)~\cite{liu2017learningl1} term in the training loss, whereas others imposed group LASSO (Wen \etal)~\cite{wen2016learning}. Ye \etal ~\cite{Ye2018RethinkingTS} similarly enforced sparsity constraint on channels during training, and the magnitude would be later used to prune the constant value channels. Lemaire \etal~\cite{2019budgetaware} introduced a budgeted regularized pruning framework that set activation volume as a budget-aware regularization term in the objective function. 
Luo \etal~\cite{luo2020residual_limited} focused on compressing residual blocks via a Kullback-Leibler (KL) divergence based criterion along with combined data augmentation and knowledge distillation to prune with limited-data. Guo \etal~\cite{guo2020dmcp} modeled channel pruning as a Markov process.  Chin \etal~\cite{chin2020towards} proposed the learning of a global ranking of the filter across different layers of the CNN. 
Gao \etal~\cite{gao2020discrete} utilized a discrete optimization method to optimize a channel-wise differentiable discrete gate with resource constraint to obtain a compact CNN with string discriminative power.

\textbf{Discussion.}
As above mentioned, pruning technique is extensive; however, those works are almost exclusively evaluated on sparse labeling classification tasks, namely, ResNet~\cite{resnet} on CIFAR-10~\cite{cifar10} and ImageNet~\cite{imagenet}. To the best of our knowledge, pruning for dense labeling networks is still a largely unexplored field. Yu \etal~\cite{yu2020joint} proposed a prune and quantize pipeline for stereo matching networks, yet their objective was different from ours. They sought extremely high sparsity for direct hardware implementation; thus, they applied \textbf{norm-based individual weight pruning} and quantized the remaining weights. The resulting model of ~\cite{yu2020joint} was \textbf{unstructured}, which required dedicated hardware to achieve the claimed ideal acceleration. Our proposed method prunes the model \textbf{structurally}, which implies that no special hardware is needed for ideal compression and acceleration. 
\section{Proposed Method} 
In this section we will present KCP, which can remove convolutional kernels whose representativeness is lower in dense labeling networks while maintaining a regular network structure.
\subsection{Preliminaries (Notation Definitions)}
We introduce notation and symbols that will be used throughout this part. We suppose that a neural network has $L$ layers. $C_{in}^j$ and $C_{out}^j$ are the number of input and output channels for the $j$th convolution layer in the network, respectively. The dimension of the $j$th convolution layer is 	$\mathbb{R}^{C_{in}^j \times C_{out}^j \times K^j \times K^j}$, where $K$ is the kernel size of the filter. Thus, in the $j$th convolution layer, there are $C_{out}^j$ filters ($\mathcal{F}$) and $C_{in}^j \times C_{out}^j$ kernels ($\mathcal{K}$). $\mathcal{K}_{m,n}^{j}$ is the $m$th kernel in the $n$th filter of $j$th layer.  Therefore, we can use kernels to represent the $j$th layer ($\mathbf{W^{(j)}}$) of the network as \{$\mathcal{K}_{m,n}^{j}$, $1\leq m \leq C_{in}^j$, and $1\leq n \leq C_{out}^j$ while $m, n \in \mathbb{N}$\}.

\subsection{Kernel Cluster Pruning} \label{sec:kcp}
Recent filter pruning methods prune at the level of channels and the target of the pruning unit is one filter.
While filter channel pruning has garnered attention in the recent research field owing to its resultant structured sparsity, removing the entire filter output channel sometimes makes it hard to identify the optimal pruning target. There are still several kernel weights in one filter. If one removes the entire filter, the interactions between each kernel in that filter are neglected and this would often result in worse accuracy after pruning. This is empirically proven in Sec.~\ref{sec:resnet}.

Neural networks designed for solving dense labeling applications often utilize different resolutions of input to obtain better prediction. Such practices leads to the need for a large receptive field in convolutional layers and requires well-designed up/downsampling layers. The performance of these networks is evaluated using pixel-wise metrics, which is prone to changes in parameters compared with typical accuracy for classification tasks. 

To solve the problem of filter pruning and dense labeling network compression while maintaining the regular network structure, we proposed KCP, a novel method that prunes \textbf{kernels} as the target instead of the filter. Our method is inspired by the concept of K-means clustering. Given a set of $n$ points as follows: $x_i \in \mathbb{R}^d, i=1,2,3,...n$ and $k$ clusters as follows: \{$S_1,S_2,...S_k \;\; k \leq n$\}. K-means aims to select the centroid that minimizes inertia between the data in the clusters and the cluster centers as follows: 
\begin{equation} \label{eq:1}
\mathop{\argmin\limits_{\mu \in \mathbb{R}^d }}\sum_{c=1}^{k}\sum_{i=1}^{n_c}{\left \| x_i-{\mu}_c  \right \|}^2\bigg|_{x_i\in S_c}
\end{equation}
$\mu_c$ denotes the cluster center and $\left \| x-\mu  \right \|$ the Euclidean distance. As the inertia measures how internally coherent the clusters are, we use the cluster center to extract the information between each kernel within each layer. In the $j$th layer, set the number of clusters $k=1$; the cluster center of that layer $\mu^j \in \mathbb{R}^{K^j \times K^j}$ is as follows:
\begin{equation} \label{eq:2}
\mu^j=\frac{1}{C_{in}^j C_{out}^j }\sum_{m=1}^{C_{in}^j}\sum_{n=1}^{C_{out}^j}\mathcal{K}_{m,n}^j 
\end{equation}
The cluster center anchors the aggregate kernel representativeness of that layer in the high dimensional space ($\mathbb{R}^{K^j \times K^j}$). If some kernels have their distances between cluster centers smaller than those of others, their representativeness are considered less significant and the information carried by them can be transferred to other kernels. Thus, these insignificant kernels can be safely pruned without hurting the network performance.
The kernel to be pruned $\mathcal{K}_{m,n}^{\prime j}$ in the $j$th layer is then determined as follows:
\begin{equation} \label{eq:3}
\argmin\limits_{m \in [1,C_{in}^j],\: n \in [1,C_{out}^j]} d\left ( m,n \right )
\end{equation}
where
\begin{equation} \label{eq:4}
d\left ( m,n \right )\: := \: \left | \mu^j - \mathcal{K}_{m,n}^j \right |
\end{equation}
is the distance of each kernel in the $j$th layer to the cluster center.
To prune more than one kernel at a time, we construct a set ($D$) from $d\left ( m,n \right )$ as follows:
\begin{equation} \label{eq:5}
D=\{x \:|\: x=d\left ( m,n \right ),\: m \in [1,C_{in}^j],\: n \in [1,C_{out}^j]\}
\end{equation}     
and calculate different quantiles ($q(p)$) of the set in Eq.~\ref{eq:5} to form a subset ($D'\subset D$):
\begin{equation} \label{eq:6}
D'=\{x'\: | \: x'\in D-\{d\left ( m,n \right )\leq q(p)\}\}
\end{equation}
$p$ denotes the portion of kernels to be removed in that layer and $q(\cdot)$ the quantile function. Kernels in $D'$ are those to be pruned. By gradually increasing the portion $p$ and iteratively pruning the kernels that are closer to the cluster center, we can achieve structured sparsity. The pruned kernels are those with insignificant representativeness and the network can easily recover its performance after fine-tuning. The complete process is shown in Algo.~\ref{algo:1}.

\begin{algorithm}[t] \label{algo:1}
	\SetAlgoLined
	\SetKw{Initialization}{Initialization:} 
	\KwIn{Network $\mathbf{W}$, target sparsity $\mathcal{S}$, pruning epochs $\mathcal{E}$, training data $\mathbf{X}$. }
	\Initialization{Pretrained parameters $\mathbf{W}=\{\mathbf{W^{(j)}},\: 1\leq j \leq L\}$,\: portion of kernels to be pruned $p=\mathcal{S}/\mathcal{E}$}\;
	\For{$epoch\leftarrow 1$ \KwTo $\mathcal{E}$}{
		Fine-tune model based on $\mathbf{X}$\;
		\For{$j\leftarrow 1$ \KwTo $L$}{
			Calculate cluster center $\mu^j$ based on Eq.~\ref{eq:2}\;
			Remove kernels in the set $D'$ of Eq.~\ref{eq:6}\;
			Update network parameter $\mathbf{W^{(j)}}$ to $\mathbf{W^{\prime(j)}}$\;
		}
		Update $p\leftarrow p+\mathcal{S}/\mathcal{E}$\;
	}
	Obtain spare network $\mathbf{W^{*}}$ from $\mathbf{W^{\prime}}$\;
	\KwOut{Compressed model with parameter $\mathbf{W^{*}}$}
	\caption{Kernel Cluster Pruning}
\end{algorithm} 
\subsection{Cost Volume Feature Distillation}
\label{sec:FD}
Knowledge Distillation (KD)~\cite{hinton2015KD} refers to the technique that helps the training process of a small \textit{student} network by having it reproduce the final output of a large \textit{teacher} network. In our work, the \textit{student} is the pruned network, whereas the \textit{teacher} is the unpruned pre-trained network. 
Unlike the conventional KD, our method guides the pruned network to mimic the prediction of some crucial high-level building blocks of the original network; thus, the resulting performance is better than directly distilling the fine-tuned output of the teacher network.

An important trait of \textbf{stereo matching} is the computation of disparity between two views. 
It is crucial to fully utilize the information within cost volume for accurate stereo matching.
We aim to transfer the 4D cost volume of the teacher to the student. 
Assuming $\mathbf{X}^{\mathcal{B}}$ to be a set of data in one mini-batch, our feature distillation (FD) loss ($\mathcal{L}_{\text{FD}}$) is expressed as follows:
\begin{equation} \label{eq:7}
\mathcal{L}_{\text{FD}} = \sum_{x_i \in \mathbf{X}^{\mathcal{B}}}KL\left (softmax(\frac{C_t}{T}), \: softmax(\frac{C_s}{T})  \right )
\end{equation} 
where $KL$ denotes the Kullback–Leibler divergence, whereas $T$ denotes the temperature. In \textbf{stereo matching}, $C_t$ and $C_s$ are the 4D cost volumes ($height \times width \times disparity\:range \times feature\:size$) generated by the teacher and student, respectively. 
The final loss function applying our feature distillation is as follows:
\begin{equation} \label{eq:8}
\mathcal{L} = \alpha \mathcal{L_{\text{task}}} + (1-\alpha)\mathcal{L_{\text{FD}}}
\end{equation}
where $\alpha$ is a constant, $\mathcal{L_{\text{task}}}$ denotes the original loss of each task, and $\mathcal{L_{\text{FD}}}$ denotes the loss described in Eq.~\ref{eq:7}. 

We would like to highlight that our focus is the novel kernel pruning technique proposed in Sec.~\ref{sec:kcp}. We merely use FD as a technique to boost the accuracy after pruning. The distillation method is exclusively applied in depth estimation. There are strong correspondences of each pixel in depth estimation (the value of label is actual estimated depth values which correlate with adjacent labels), whereas in segmentation, such a connection is less significant (the value of label is merely encoding of different object classes). This indicates that our high level feature distillation method would perform better in transferring cost volume in depth estimation because cost volume carries the important correspondence information. Suppose that one is interested in segmentation distillation results, we also provide those in the supplementary materials.  
\section{Experiments}
We first benchmark our KCP on ResNet~\cite{resnet} trained on CIFAR-10~\cite{cifar10} and ILSVRC-2012~\cite{imagenet} to verify the effectiveness of our method. Subsequently, for dense labeling applications, the experimental evaluations are performed on end-to-end networks. For stereo matching, we select PSMNet~\cite{PSMNet} trained on KITTI~\cite{KITTI2015C, KITTI2015J}. HRNet~\cite{HRNet} has been the backbone of several SOTA semantic segmentation networks. We test our pruning method on HRNet trained on Cityscapes~\cite{Cordts2016Cityscapes} 
to verify the effect of pruning on semantic segmentation. Feature distillation is applied on PSMNet.
\subsection{Experimental Settings}
\label{sec:exp_setting}
\textbf{Architectures and Metrics.}
For ResNet, we use the default parameter settings of ~\cite{resnet}. Data augmentation strategies are the same as in PyTorch~\cite{pytorch}’s official examples. The pre-trained models are from official torchvision models. We report Top-1 accuracy.

For PSMNet, the pre-trained model and implementation are from original authors. The error metric is the 3-px error, which defines error pixels as those having errors of more than
three pixels or 5\% of disparity error. The 3-px accuracy is simply the 3-px error subtracted from $100\%$. We calculate the 3-px error directly using the official code implementation of the author. We prune for 300 epochs and follow the training setting of~\cite{PSMNet}.

For HRNet, we use the "HRNetV2-W48" configuration with no multi-scale and flipping. The implementation and training settings follows~\cite{sun2019highresolution}. Following the convention of semantic segmentation, the mean of class-wise intersection over union (mIoU) is adopted as the evaluation metric.

\textbf{Datasets.}
The CIFAR-10 dataset contains 50,000 training images and 10,000 testing images. The images are $32\times 32$ color images in 10 classes. ILSVRC-2012 contains 1.28 million training images and 50 k validation images of 1,000 classes.

The KITTI2015 stereo is a real world dataset of street views. It consists of 200 training stereo color image pairs with sparse ground-truth disparities obtained from LiDAR and 200 testing images. The training images are further divided into training set (160) and validation set (40). Image size is $376\times 1240$.

The Cityscapes dataset contains 5,000 finely annotated color images of street views. The images are divided as follows: 2,975 for training, 500 for validation, and 1,525 for testing. There are 30 classes, but only 19 classes are used for evaluation. Image size is $1024\times 2048$.

\subsection{KCP on Classification} \label{sec:resnet}
\begin{table}[h!]
	\centering
	\scriptsize
	\caption{Pruning results of ResNet-56/110 on CIFAR-10. The Acc. $\downarrow$ is the accuracy drop between pruned model and baseline. Negative Acc. $\downarrow$ indicates that the accuracy is better than the baseline.}
	\label{table:cifar}
	\begin{tabular}{|c|c|c|c|c|c|c|}
		\hline
		Depth                & Method & Baseline & Acc.  & Acc.$\downarrow$ & Sparsity & FLOPs    \\ \hline \hline
		\multirow{6}{*}{56}  & PFEC~\cite{li2016pruningfilter}   & 93.04    & 91.31 & 1.73  & 13.70\%  & 9.09E7 \\  
		& FPGM~\cite{he2019GM}   & 93.59    & 92.93 & 0.66  & 38.71\%  & 5.94E7 \\  
		& LPFC~\cite{he2020LFPC}   & 93.59    & 93.34 & 0.25  &     -     & 5.91E7 \\  
		& Ours   & 93.69    & 93.37 & 0.32  & 39.58\%  & \textbf{5.27E7} \\ 
		& FPGM~\cite{he2019GM}   & 93.59    & 92.04 & 1.55  & 59.13\%  & 3.53E7         \\  
		& Ours   & 93.69    & 93.31 & \textbf{0.38}  & \textbf{60.54\%}  & \textbf{2.76E7} \\ \hline \hline
		\multirow{6}{*}{110} & PFEC~\cite{li2016pruningfilter}   & 93.53    & 92.94 & 0.59  & 32.40\%  & 1.55E8 \\ 
		& FPGM~\cite{he2019GM}   & 93.68    & 93.73 & -0.05 & 38.73\%  & 1.21E8 \\  
		& LPFC~\cite{he2020LFPC}   & 93.68    & 93.79 & -0.11 &    -      & 1.01E8 \\  
		& Ours   & 93.87    & 93.79 & 0.08  & 50.74\%  & \textbf{9.20E7} \\  
		& FPGM~\cite{he2019GM}   & 93.68    & 93.24 & 0.44  & 59.16\%  & 7.19E7         \\  
		& Ours   & 93.87    & 93.66 & \textbf{0.21}  & \textbf{59.64\%}  & \textbf{7.10E7} \\ \hline
	\end{tabular}
\end{table}

\begin{table*}[h!]
	\centering
	\small
	\caption{Pruning results of ResNet on ILSVRC-2012. The $\downarrow$ has the same meaning as in Table~\ref{table:cifar}.}
	\label{table:imagenet}
	\begin{tabular}{|c|c|c|c|c|c|c|c|c|c|}
		\hline
		Depth        & Method  & {Baseline Top1/ 5} & Top1(\%)   & Top1$\downarrow$(\%) & Top5(\%)   & Top5$\downarrow$(\%) & Sparsity & FLOPs & FLOPs$\downarrow$ \\ \hline \hline
		\multirow{5}{*}{18} & FPGM~\cite{he2019GM}   & 70.28 / 89.63             & 68.41  & 1.87    & 88.48  & 1.15    & 28.10\%  & 1.08E9 & 41.8\%\\
		& DMCP~\cite{guo2020dmcp}   & 70.10 / -             & 69.20  & 0.9    & -  & -    & -  & 1.04E9 & 43.8\%\\
		&  Ours       & 69.76 / 89.08             & 70.10  & \textbf{-0.34}    & 89.45   & \textbf{-0.33}   & 27.2\%  & 1.06E9 & 42.8\%\\
		& FPGM~\cite{he2019GM}   & 70.28 / 89.63             & 62.23  & 8.05    & 84.31  & 5.32    & 56.34\%  & 5.53E8 & 70.2\%\\ 
		& Ours            & 69.76 / 89.08             & 63.79  & \textbf{5.97}   & 85.56  & \textbf{3.52} & 55.86\% & \textbf{4.70E8} & \textbf{75.1\%}\\\hline \hline
		\multirow{6}{*}{34} & PFEC~\cite{li2016pruningfilter}                  & 73.23 / -                 & 72.17  & 1.06    & -      & -       & 10.80\%  & 2.76E9 & 24.2\%\\
		& FPGM~\cite{he2019GM}          & 73.92 / 91.62             & 72.63  & 1.29    & 91.08  & 0.54    & 28.89\%  & 2.18E9 & 41.1\%\\
		& DMC~\cite{gao2020discrete}     & 73.30 / 91.42                 & 72.57  & 0.73    & 91.11      & 0.31       & -  & 2.10E9 & 43.4\%\\
		& Ours       & 73.31 / 91.42             & 73.46 & \textbf{-0.15}   & 91.56  & \textbf{-0.14}    & \textbf{29.25\%}  & \textbf{2.04E9} & \textbf{44.9\%}\\
		& FPGM~\cite{he2019GM}          & 73.92 / 91.62             & 67.69  & 6.23    & 87.97  & 3.65    & 57.94\%  & 1.08E9 & 70.9\%\\
		& Ours              & 73.31 / 91.42             & 69.19 & \textbf{4.12}   & 88.89 &  \textbf{2.53}   & 57.93\%  & \textbf{9.19E8} & \textbf{75.2\%}\\ \hline \hline
		\multirow{10}{*}{50} 
		& DMCP~\cite{guo2020dmcp}   & 76.60 / -             & 77.00  & -0.40    & -  & -    & -  & 2.80E9 & 27.4\%\\
		& SFP~\cite{he2018SoftFP}           & 76.15 / 92.87             & 62.14 &  14.01  & 84.60 & 8.27 & 30.00\%  & 2.24E9 & 41.8\%\\
		& HRank~\cite{lin2020hrank}           & 76.15 / 92.87             & 74.98 &  1.17  & 92.33 & 0.51 & 36.67\%  & 2.30E9 & 43.8\%\\
		& FPGM~\cite{he2019GM}          & 76.15 / 92.87             & 75.59  & 0.56    & 92.23  & 0.24    & 24.26\%  & 2.23E9 & 42.2\%\\
		& Ours       & 76.13 / 92.87             & 76.61 & \textbf{-0.48}   & 93.21  & \textbf{-0.34}    & 31.19\%  & \textbf{2.18E9} & 43.5\%\\
		& FPGM~\cite{he2019GM}          & 76.15 / 92.87             & 74.83  & 1.32    & 92.32  & 0.55    & 32.37\%  & 1.79E9 & 53.5\%\\
		& DMC~\cite{gao2020discrete}     & 76.15 / 92.87                 & 75.35  & 0.80    & 92.49      & 0.38       & -  & 1.74E9 & 55.0\%\\
		& Ours              & 76.13 / 92.87             & 76.26 & \textbf{-0.13}   & 93.04  & \textbf{-0.17}    & \textbf{37.06\%}  & \textbf{1.77E9} & \textbf{54.1\%}\\
		& LFPC~\cite{he2020LFPC}              & 76.15 / 92.87             & 74.46 & 1.69   & 92.04  & 0.83    & -  & 1.51E9 & 60.8\%\\ 
		& Ours               & 76.13 / 92.87             & 74.80 & \textbf{1.33}   & 92.19  &  \textbf{0.68}   & 46.11\%  & \textbf{1.30E9} & \textbf{66.3\%}\\\hline \hline
		\multirow{2}{*}{101} & FPGM~\cite{he2019GM}     & 77.37 / 93.56                 & 77.32  & 0.05    & 93.56      & 0.00       & 26.63\%  & 4.37E9 & 42.2\%\\
		& Ours     & 77.37 / 93.56                 & 77.92  & \textbf{-0.55}    & 93.92      & \textbf{-0.36}       & 28.48\%  & 4.41E9 & 41.7\% \\ \hline
	\end{tabular}
\end{table*}

We compare our method with other filter-level pruning works. For the CIFAR-10 dataset, we report accuracy on ResNet-56 and 110 trained from scratch. The pruning epoch $\mathcal{E}$ is set to 200 and no additional fine-tune epochs are added. As shown in Table~\ref{table:cifar}, our method achieves SOTA results. Our method can achieve comparable accuracy with previous works at low sparsity, whereas at high sparsity, our method can provide more sparsity (FLOPs reduction) with less accuracy drop.

For the ILSVRC-2012 dataset, we test pre-trained ResNet-18, 34, 50, and 101. The pruning epoch $\mathcal{E}$ is set to 100. Table~\ref{table:imagenet} shows that our method outperforms those of other works. For approximately 30\% sparsity, our method can even achieve accuracy higher than the baseline. At higher sparsity, KCP also achieves better accuracy with larger FLOPs reduction. For example, for ResNet-50, our method achieves similar FLOPs reduction with FPGM~\cite{he2019GM} and DMC~\cite{gao2020discrete}, but our model exceeds theirs by 1.45\% and 0.93\% on the accuracy, respectively (See row 8 of ResNet-50 in Table~\ref{table:imagenet}). Moreover, for pruning pretrained ResNet-101, our method can reduce 41.7\% of FLOPs and even gain 0.55\% accuracy. A similar observation can also be found on ResNet-18 and ResNet-34. KCP utilizes a smaller target for pruning and considers the relationship between different kernels, which is the reason for its success over other previous works. Proved by all these results, our method achieves SOTA results.

\subsection{KCP on Depth Estimation}
Recent works have shown that depth estimation from stereo pair images can be formulated as a supervised learning task resolved by neural networks. Successful depth estimation cannot be achieved without exploiting context information to find correspondence. The use of stacked 3D convolution can refine the 4D cost volume and thus provide better results. However, the 3D convolution inevitably increases the parameter and computation complexity, which enables our kernel pruning method to leverage the redundancy.
Table~\ref{table:psmnet} summarizes the pruning result of PSMNet~\cite{PSMNet} accuracy drop under different sparsity configurations. Our method can achieve 77\% of FLOPs reduction with less than 1\% of accuracy drop (P. 60.13\%) and can even reduce more than 90\% of FLOPs with only 3\% of accuracy loss (P. 81.95\%).

\begin{figure}[h!]
	\centering
	\includegraphics[scale=0.23]{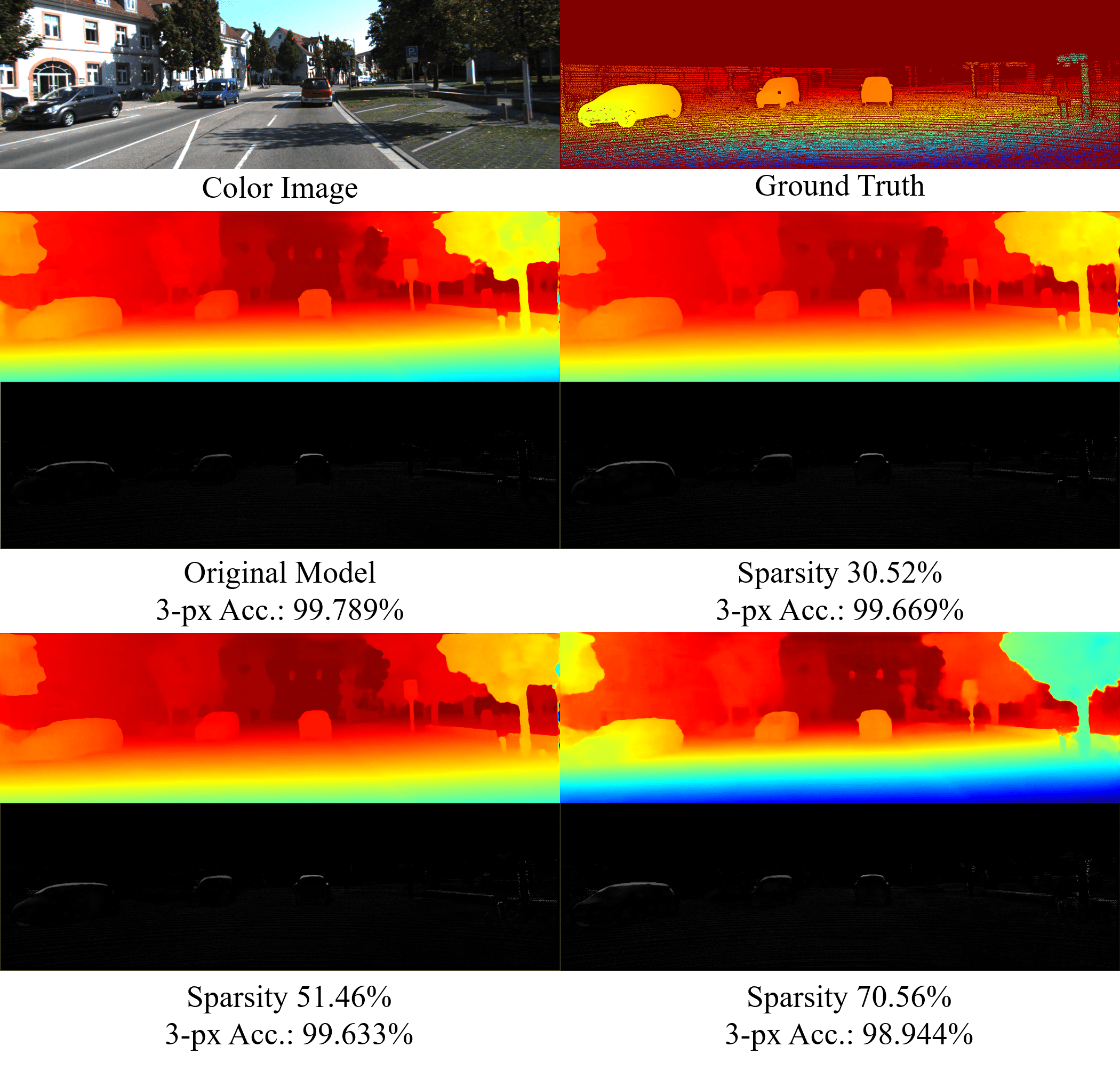}
	\caption{Results of our method of PSMNet on KITTI2015. We show the predicted depth map and the difference with ground truth (brightness indicates difference) from unpruned original model, 30\% sparsity, 50\% sparsity and 70\% sparsity. The 3-px accuracies are listed below each predicted result. The model performance is well preserved both qualitatively and quantitatively.}
	\label{fig:psmnet_vis}
\end{figure}

The predicted depth maps from KITTI2015 dataset of different sparsity models are displayed in Fig.~\ref{fig:psmnet_vis}. We also provide the difference map between ground truth and predictions. The darker region indicates more difference. Because the ground truths of the KITTI2015 dataset are sparse, we follow the convention by only considering the valid points during the difference map generation. Notably, the depth maps predicted by pruned models are almost identical to those of the unpruned original model. Only some edge areas are predicted differently from the ground truth.
These results demonstrate that our method can effectively compress the stereo matching model with negligible performance loss.  
\begin{table}[h!]
	\centering
	\small
	\caption{Accuracy and FLOPs of different sparsity configuration of PSMNet on KITTI2015. P.30.52\% means that the model is pruned to 30.52\% sparsity. The FLOPs are calculated under the assumption that the network runs at full HD resolution and 60 fps.}
	\label{table:psmnet}
	\begin{tabular}{ccccc}
		\toprule
		& 3-px Acc. & 3-px Acc.↓ & FLOPs    & FLOPs ↓ \\ \midrule
		Original       & 99.10\%   &  -          & 2.43E14 &  -       \\ \midrule
		P. 30.52\% & 98.81\%   & 0.29\%     & 1.32E14 & 45.68\% \\ \midrule
		P. 51.46\% & 98.56\%   & 0.54\%     & 7.47E13 & 69.26\% \\ \midrule
		P. 60.13\% & 98.50\%   & 0.61\%     & 5.54E13 & 77.20\% \\ \midrule
		P. 70.56\% & 97.59\%   & 1.52\%     & 3.57E13 & 85.31\% \\ \midrule
		P. 81.95\% & 96.07\%   & 3.06\%     & 1.85E13 & 92.39\% \\ \bottomrule
	\end{tabular}
\end{table}  

\subsection{KCP on Semantic Segmentation}
\begin{table}[h!]
	\centering
	\small
	\caption{Mean IoU and FLOPs of different sparsity of HRNet on Cityscapes. Negative meanIoU$\downarrow$ means that the meanIou is higher than the original unpruned model. P.20.35\% denotes that the model is pruned to 20.35\% sparsity. The FLOPs are calculated with the network taking an input of size $512\times 1024$.}
	\label{table:hrnet}
	\begin{tabular}{ccccc}
		\toprule
		& mIoU & mIoU↓(\%) & FLOPs & FLOPs↓(\%)  \\ \midrule
		Original       & 0.8090    &     -          & 1.74E11& -\\ \midrule
		P. 20.35\% & 0.8121   & \textbf{-0.383}        &  1.13E11        &35.06\%\\ \midrule
		P. 29.95\% & 0.8096   & \textbf{-0.074}        &  8.78E10        &49.54\%\\ \midrule
		P. 50.31\% & 0.8040   & 0.618         &  4.39E10        &74.77\%\\ \midrule
		P. 59.90\% & 0.7924   & 2.052         &  2.74E10        &84.25\%\\ \midrule
		P. 69.49\% & 0.7873   & 2.682         &  1.35E10        &92.24\%\\ \bottomrule
	\end{tabular}
\end{table}
\begin{figure}[h!]
	\centering
	\includegraphics[scale=0.575]{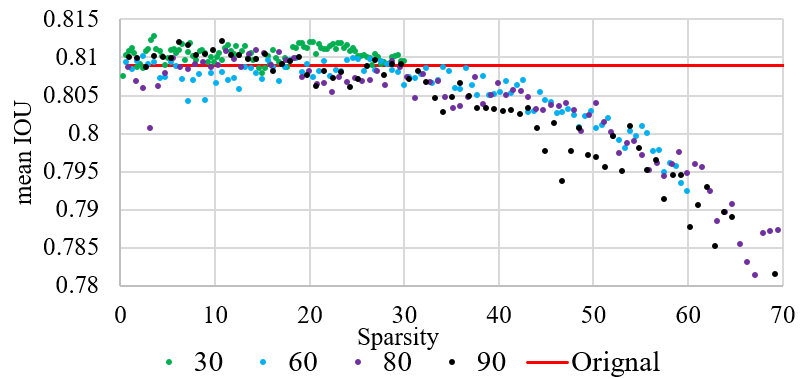}
	\caption{Validation mean IoU of HRNet pruned with different $\mathcal{S}$.}
	\label{fig:hrnet_spar}
\end{figure}
\begin{figure}[h!]
	\centering
	\includegraphics[scale=0.2345]{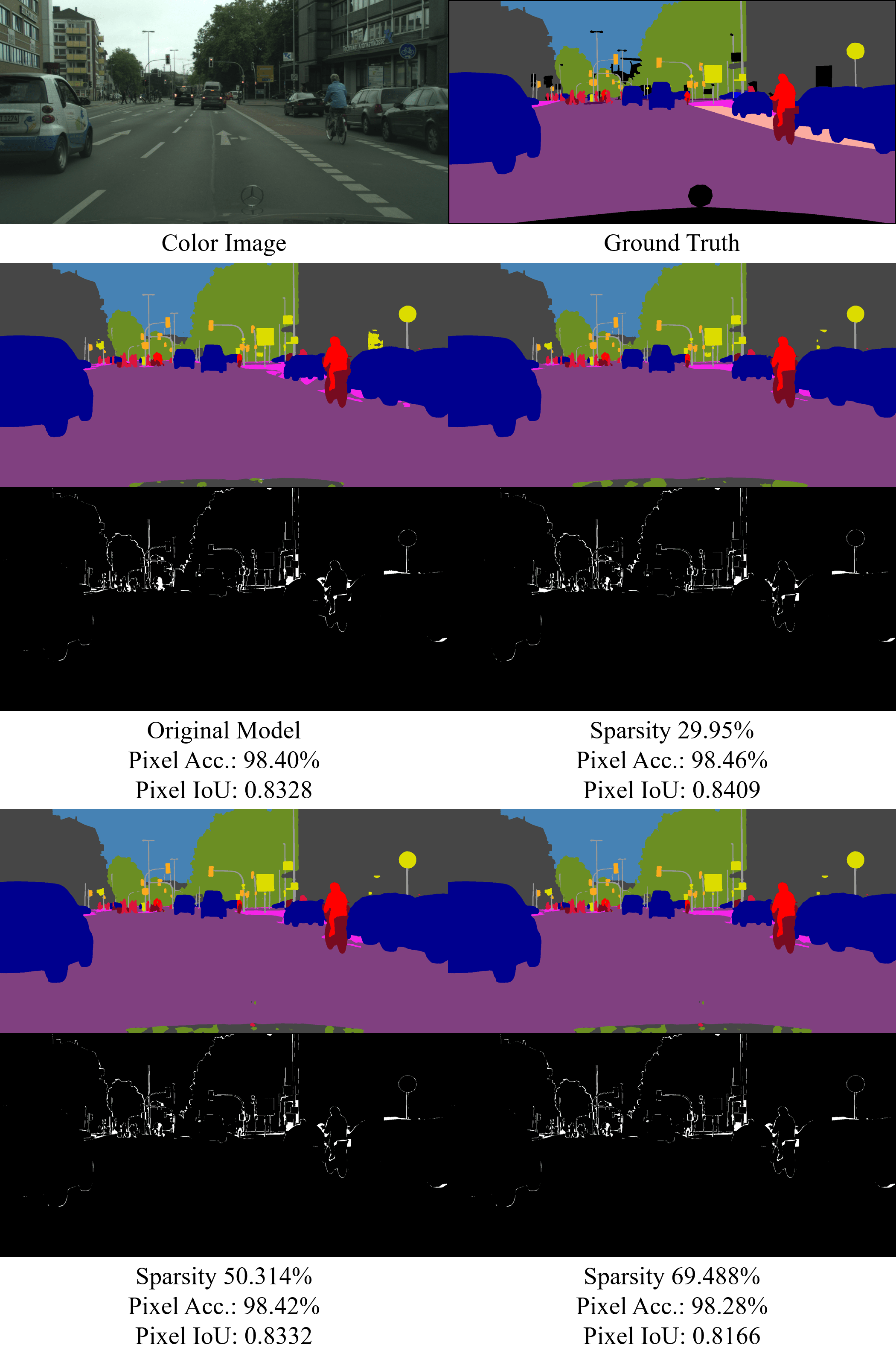}
	\caption{Results of our method of HRNet on Cityscapes. The predicted class label and the difference between ground truth are shown, (brightness indicates difference) including original model, 30\% sparsity, 50\% sparsity and 70\% sparsity.  The pixel accuracy and pixel IoU are listed below each figure.}
	\label{fig:hrnet_vis}
\end{figure}
High-Resolution network augments the high-resolution representations output from different stages by aggregating the (upsampled) representations from all the parallel convolutions. The different resolution blocks comprise convolutional layers with a large receptive field, which contains significant redundancy for our kernel pruning method to optimize. Fig.~\ref{fig:hrnet_spar} illustrates the change of validation mean IoU of HRNet~\cite{HRNet} pruned with different $\mathcal{S}.$ Notably, at sparsity of less than 30\%, the performance of the pruned model can even exceed that of the original model. The mean of IoU degradation is within 2.7\% even at 70\% sparsity. Table~\ref{table:hrnet} summarizes the mIoU drop and resulting FLOPs of the pruned models. Based on the result of HRNet pruning, our method is proven to be effective on semantic segmentation, again\; we can reduce 35\% of FLOPs with 0.38\% of performance gain, and can also reduce 74\% of FLOPs with less than 1\% of performance loss. The visualization results of the predicted class label of Cityscapes validation images are shown in Fig.~\ref{fig:hrnet_vis}. To highlight the difference, we calculate the difference map between predicted class labels and ground truth on the 19 classes used for evaluation. The black region indicates identical class labels, whereas the white region indicates different class labels with ground truth. Other classes appearing in the ground truth, but are ignored in evaluation are not shown on the difference map. The difference maps of the pruned models in Fig.~\ref{fig:hrnet_vis} are almost identical to those of the original model. The results demonstrate that our method can remove redundant weights without hurting the network performance. 

\subsection{Ablation Study}
We conduct experiments to perform ablation studies in this section. The effect of the proposed cost volume feature distillation will be presented. The adversarial criteria is also added to further validate the effectiveness of our method.

\begin{figure}[h!]
	\centering
	\includegraphics[scale=0.627]{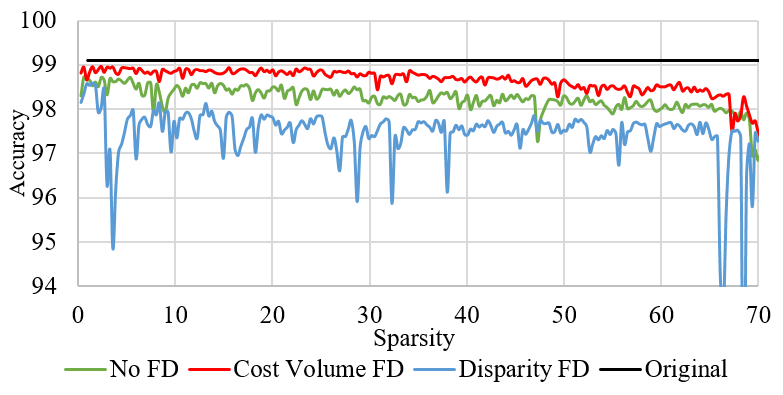}
	\caption{Variants of feature distillation for PSMNet pruning.}
	\label{fig:ablation_psm_kd}
\end{figure}

\textbf{Feature Distillation.} Variants of feature distillation are applied to demonstrate the appropriate distillation target for increasing model performance. Corresponding results of PSMNet are presented in Fig.~\ref{fig:ablation_psm_kd}. For the network being pruned, "No FD" does not apply feature distillation, "Cost Volume FD" utilizes 4D cost volume as the distillation target as described in Eq.~\ref{eq:7}, and "Disparity FD" uses final disparity output as the target for the conventional knowledge distillation. We set $\alpha$ to 0.9 and $T$ to 15. Results shown in Fig.~\ref{fig:ablation_psm_kd} well demonstrate that by encouraging the pruned network to mimic the 4D cost volume construction of the original network (Cost Volume FD), the accuracy after pruning can be better maintained. Distilling the final output even results in worse accuracy than not applying distillation. We believe that the correct choice of crucial feature is the key to better model performance. Because the construction of cost volume plays a critical role in stereo matching, distilling the feature of cost volume provides the best results.   


\begin{table}[h!]
	\centering
	\caption{Analysis of adversarial criteria. Models are ResNet-56 and ResNet-110. $\mathcal{S}$ is set to 30. 10\% Adv. denotes that 10\% of the layers are randomly chosen to be pruned using adversarial criteria.}
	\label{table:adv_2}
	\begin{tabular}{ccccc}
		\toprule
		& \multicolumn{2}{c}{ResNet-56} & \multicolumn{2}{c}{ResNet-110} \\ \midrule
		Config.   & Acc.       & Sparsity      & Acc.       & Sparsity       \\ \midrule
		Non Adv.  & 93.62\%        & 29.69\%       & 94.05\%        & 29.80\%        \\ \midrule
		10\% Adv. & 74.85\%        & 30.14\%       & 76.88\%        & 29.47\%        \\ \midrule
		20\% Adv. & 51.96\%        & 28.13\%       & 36.75\%        & 26.99\%        \\ \midrule
		40\% Adv. & 14.35\%        & 29.72\%       & 13.28\%        & 27.23\%        \\ \bottomrule
	\end{tabular}
\end{table}

\textbf{Adversarial Criteria.} We apply adversarial criteria to further validate the effectiveness of our method. Adversarial criteria means that in contrast to Eq.~\ref{eq:3} in which we prune the kernels that are the \textbf{closest} to the cluster center, we instead remove the kernels that are the \textbf{farthest} from the cluster center. Table~\ref{table:adv_2} shows that even only replacing 10\% of our method into adversarial criteria significantly degrades the accuracy (10\% Adv.). The accuracy becomes worse when the percentage of adversarial criteria increases, indicating that our method can truly select the kernels that are the least representational in each layer; evidence of the effectiveness of our proposed method.  
\subsection{Further Exploration}
\begin{table}[h!] 
	\centering
	\caption{Mean PSNR and SSIM of unpruned and pruned SRGAN generator on recovering DIV2K validation images.}
	\label{table:srgan}
	\begin{tabular}{ccc}
		\toprule
		& PSNR (dB)    & SSIM     \\ \midrule
		Finetune      & 28.1009 & 0.7968  \\ \midrule
		Prune 37.16\% & 27.6507 & 0.7937 \\ \midrule
		Prune 53.16\% & 27.7121 & 0.7887 \\ \bottomrule
	\end{tabular}
\end{table}
\begin{figure}[h!]
	\centering
	\includegraphics[scale=0.215]{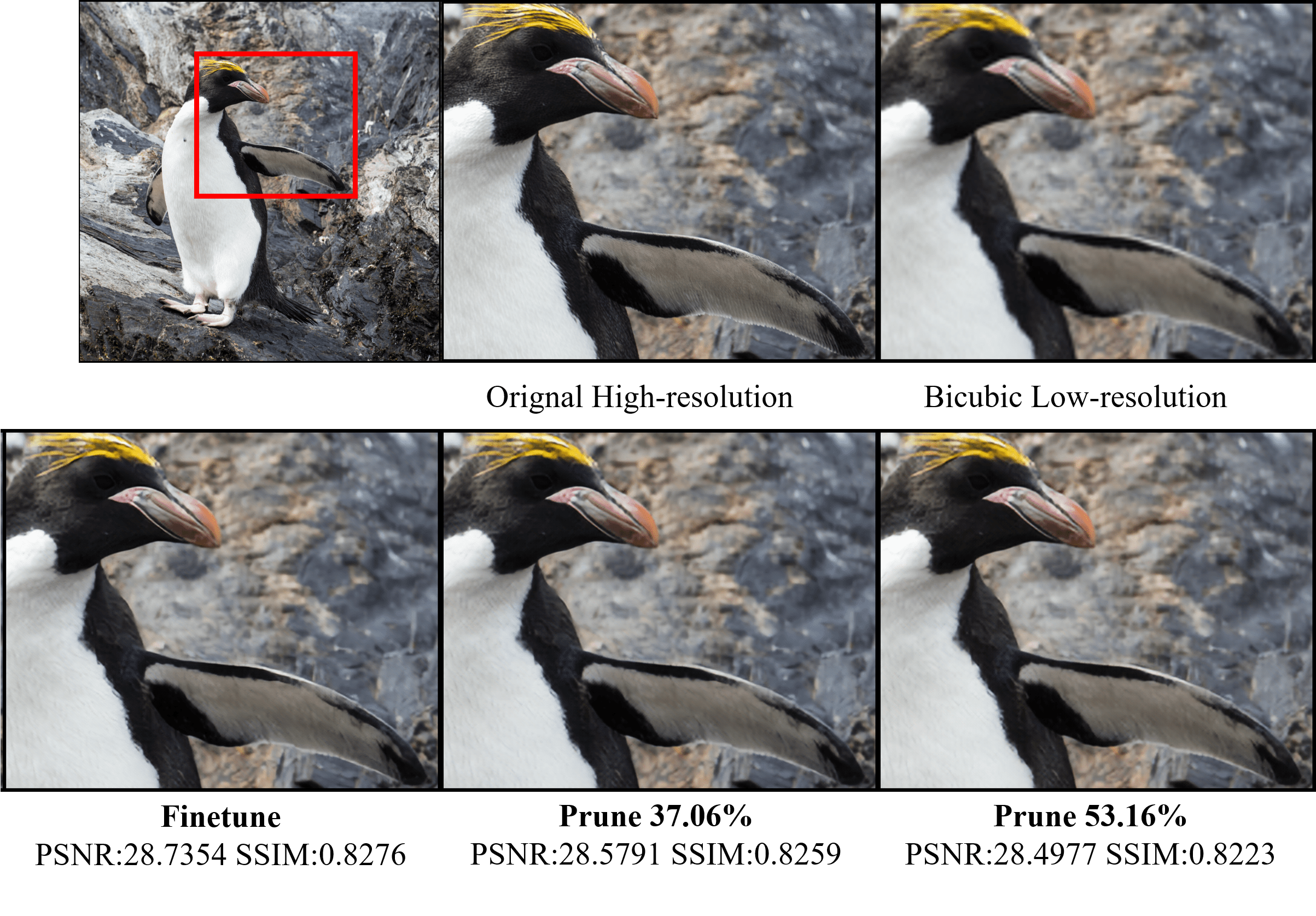}
	\caption{Corresponding high resolution image, bicubic-upsampled low resolution image, and reconstruction results from different sparsity SRGAN generators. The image is from DIV2K validation set. Results are zoomed-in from the red block.}
	\label{fig:srgan1}
\end{figure}
To further validate our KCP, we apply pruning on SRGAN~\cite{SRGAN}, which recovers a single image resolution using a generative adversarial network (GAN). We are curious about how pruning affects the image generation of GAN and the perceptual difference induced by pruning on restored high-resolution images. 

The SRGAN is first trained on color images sampled from VOC2012~\cite{everingham2015pascal} dataset for 100 epochs and then fine-tuned on DIV2K~\cite{DIV2K} dataset. We only prune the generator and keep the discriminator updated, because during inference, only the generator is used to recover low resolution images. The experiments are performed with a scale factor of $4 \times$ between low- and high-resolution
images. We report two classical super-resolution performance measurements PSNR (dB) and SSIM for comparison.

Table~\ref{table:srgan} shows the results. The pruned generator can achieve similar PSNR and SSIM to the unpruned model. In terms of human perception, Fig.~\ref{fig:srgan1} demonstrates that the pruned generator still reconstructs the details quite well. Little difference can be observed by human perception. 
This means that our KCP successfully selected the low representative kernels and the success on GAN pruning paves the way for explainable pruning. In the future, the interaction between specific parameters and human perception may be observed through more experiments on GAN pruning.
  
\section{Conclusions}
In this paper, we present a novel structured pruning method KCP, which explores the representativeness of parameters by clustering filter kernels, to elucidate dense labeling neural network pruning. We demonstrate that by removing the kernels that are closest to the cluster center in each layer, we can achieve high FLOPs reduction with negligible performance loss on dense labeling application models. In addition, we propose cost volume feature distillation to boost the accuracy of depth estimation network after pruning. Extensive experiments on various CNNs demonstrate the effectiveness of KCP in reducing model size and computation overhead while showing little compromise on the model performance. This paves the way for efficient model deployment of those real-time dense labeling algorithms. Notably, KCP achieves new SOTA results for ResNet on both CIFAR-10 and ImageNet datasets. We also explore the perception difference induced by pruning through applying KCP on GAN. We believe that in the future, generative model pruning may help researchers to develop human perception based pruning.   

{\small
\bibliographystyle{ieee_fullname}
\bibliography{egbib}
}

\clearpage
\appendix
\section{ResNet on CIFAR-100}
CIFAR-100 is a dataset just like the CIFAR-10~\cite{cifar10}, except that it has 100 classes containing 600 images each. There are 500 training images and 100 testing images per class. The experimental setting of ResNet on CIFAR-100 follows the setting mentioned in Sec.~\ref{sec:exp_setting}. Since there are few methods having experiments on CIFAR-100, we only list LFPC~\cite{he2020LFPC} for comparison. The result of pruning ResNet-56 on CIFAR-100 is presented in Table~\ref{table:cifar100_comp}. When achieving slightly better FLOPs reduction, our KCP can result in better accuracy after pruning. In fact, the accuracy of KCP-pruned ResNet-56 even increases by 0.23\%. This further proves the effectiveness of our method.

Additional pruning results for different depth ResNet on CIFAR-100 are listed in Table~\ref{table:cifar100}. 
\begin{table}[h!]
	\centering
	\caption{Comparison of pruning ResNet-56 on CIFAR-100.}
	\label{table:cifar100_comp}
	\begin{tabular}{ccccc}
		\toprule
		Depth               & Method & Acc.↓(\%) & FLOPs    & FLOPs ↓ \\ \cmidrule{1-5}
		\multirow{2}{*}{56} & LFPC~\cite{he2020LFPC}   & 0.58   & 6.08E+07 & 51.60\% \\ \cmidrule{2-5} 
		& Ours   & -0.23  & 5.95E+07 & 52.57\% \\ \bottomrule
	\end{tabular}
\end{table}

\begin{table}[h!]
	\centering
	\caption{Results of KCP processed ResNet on CIFAR-100.}
	\label{table:cifar100}
	\begin{tabular}{ccccc}
		\toprule
		& \multicolumn{4}{c}{Baseline (Original   Unpruned Models)}  \\ \midrule
		Depth                & \multicolumn{2}{c}{Acc.}   & \multicolumn{2}{c}{FLOPs}    \\ \midrule
		20                   & \multicolumn{2}{c}{68.23}  & \multicolumn{2}{c}{4.06E+07} \\ \midrule
		32                   & \multicolumn{2}{c}{69.62}  & \multicolumn{2}{c}{6.89E+07} \\ \midrule
		56                   & \multicolumn{2}{c}{71.27}  & \multicolumn{2}{c}{1.25E+08} \\ \midrule
		86                   & \multicolumn{2}{c}{72.76}  & \multicolumn{2}{c}{1.96E+08} \\ \midrule
		110                  & \multicolumn{2}{c}{73.31}  & \multicolumn{2}{c}{2.53E+08} \\ \bottomrule \toprule
		& \multicolumn{4}{c}{KCP Pruned Models}                      \\ \hline
		Depth                & Acc. ↓(\%)      & Sparsity      & FLOPs          & FLOPs ↓      \\ \hline
		\multirow{3}{*}{20}  & 1.07        & 34.20\%       & 1.98E+07       & 51.22\%      \\ \cline{2-5} 
		& 2.49        & 38.66\%       & 1.76E+07       & 56.60\%      \\ \cline{2-5} 
		& 3.45        & 51.58\%       & 1.20E+07       & 70.47\%      \\  \hline \hline
		\multirow{3}{*}{32}  & -0.06       & 34.76\%       & 3.32E+07       & 51.86\%      \\ \cline{2-5} 
		& 0.89        & 43.42\%       & 2.62E+07       & 86.64\%      \\ \cline{2-5} 
		& 1.96        & 52.10\%       & 2.00E+07       & 70.91\%      \\  \hline \hline
		\multirow{5}{*}{56}  & -0.32       & 26.23\%       & 7.43E+07       & 40.80\%      \\ \cline{2-5} 
		& -0.23       & 35.38\%       & 5.95E+07       & 52.57\%      \\ \cline{2-5} 
		& 0.42        & 43.96\%       & 4.71E+07       & 62.49\%      \\ \cline{2-5} 
		& 0.89        & 53.33\%       & 3.51E+07       & 72.05\%      \\ \cline{2-5} 
		& 2.89        & 60.84\%       & 2.67E+07       & 78.76\%      \\ \hline \hline
		\multirow{5}{*}{86}  & -0.29       & 27.49\%       & 1.13E+08       & 42.36\%      \\ \cline{2-5} 
		& 0.65        & 36.44\%       & 9.09E+07       & 53.71\%      \\ \cline{2-5} 
		& 0.76        & 44.08\%       & 7.37E+07       & 62.46\%      \\ \cline{2-5} 
		& 1.52        & 52.89\%       & 5.60E+07       & 71.46\%      \\ \cline{2-5} 
		& 1.21        & 61.32\%       & 4.12E+07       & 79.00\%      \\ \hline \hline
		\multirow{5}{*}{110} & 0.43        & 28.43\%       & 1.73E+08       & 31.75\%      \\ \cline{2-5} 
		& 0.64        & 35.52\%       & 1.20E+08       & 52.63\%      \\ \cline{2-5} 
		& 1.06        & 45.37\%       & 9.14E+07       & 63.86\%      \\ \cline{2-5} 
		& 1.44        & 52.36\%       & 7.34E+07       & 70.97\%      \\ \cline{2-5} 
		& 1.92        & 61.08\%       & 5.36E+07       & 78.82\%      \\ \hline
	\end{tabular}
\end{table}
\section{More Ablation Study}
\begin{figure}[h!]
	\centering
	\includegraphics[scale=0.4]{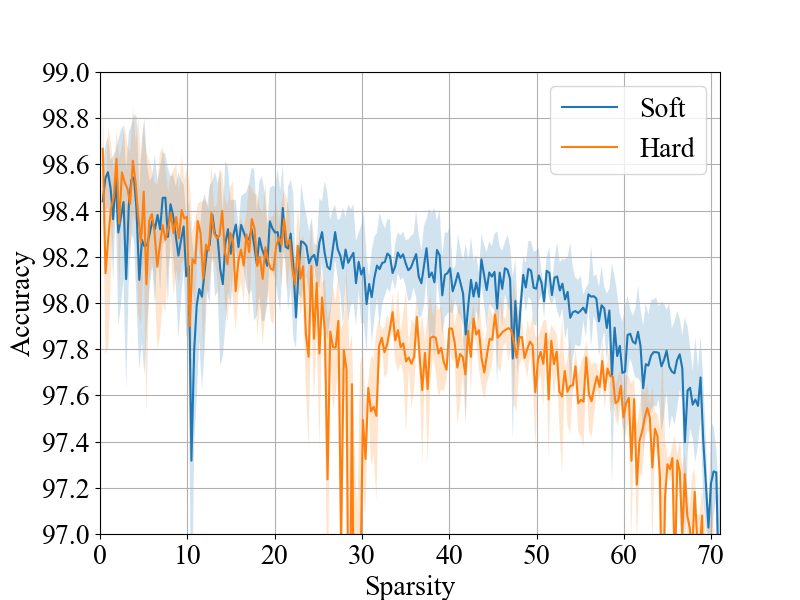}
	\caption{The influence of different fine-tuning strategy. Line represents the average over 3 experiment and the shaded area are the upper and lower bound of the standard deviation.}
	\label{fig:ablation_psm_finetune}
\end{figure}	

\begin{figure}[h!]
	\centering
	\includegraphics[scale=0.4]{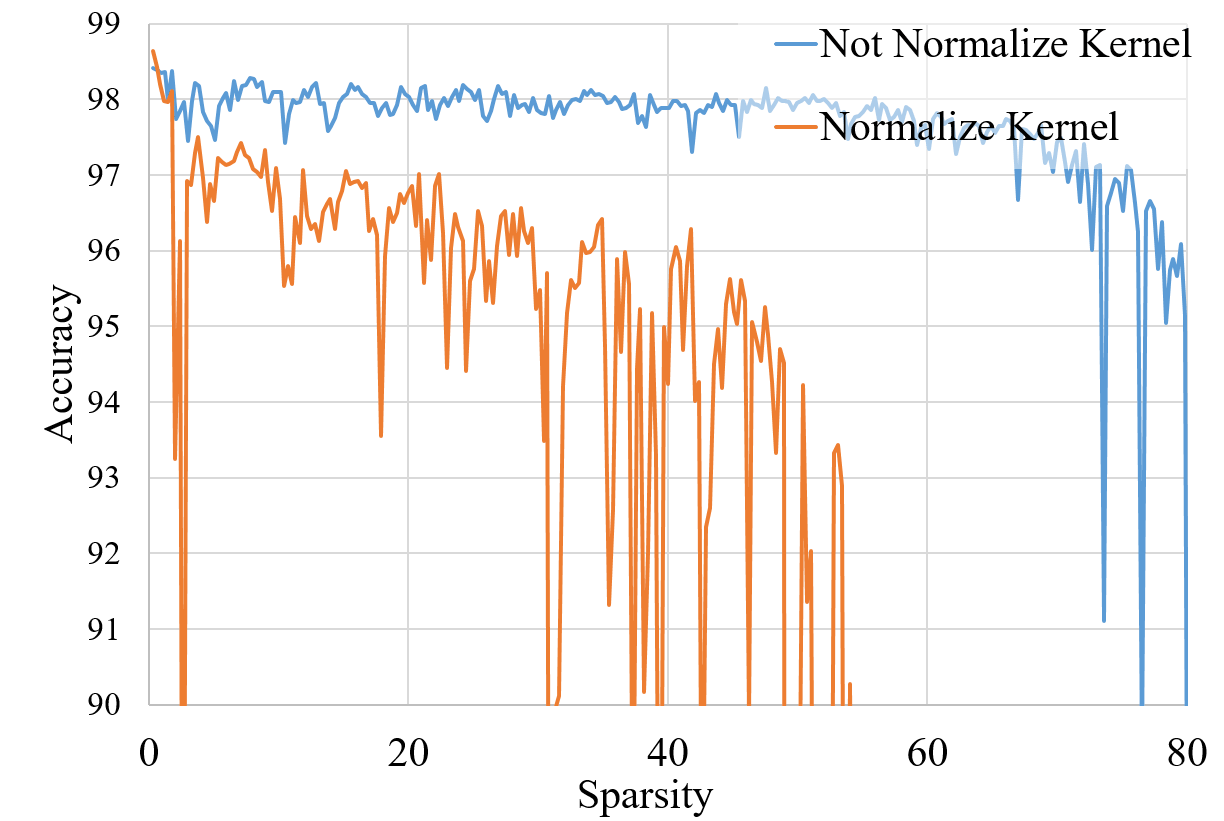}
	\caption{The influence of normalizing the kernel in each layer before performing clustering.}
	\label{fig:ablation_psm_normalize}
\end{figure}

\textbf{Fine-tune Strategy and Kernel Normalization.} For brevity, we report the ablative results of the PSMNet on these two variant categories. Similar observation can be found in other network and dataset that we used in our work. Two common fine-tune strategies, soft and hard, are examined with our kernel cluster pruning. Soft means the removed kernel can be recovered with updated value throughout the pruning process, while hard means the removed weight remains zeros whenever it is pruned. As previous mentioned in Sec~\ref{sec:kcp}, we adopted soft fine-tuning in our method. Fig.~\ref{fig:ablation_psm_finetune} validates that soft fine-tuning stabilizes the pruning process and results in better accuracy. Soft fine-tuning recovers pruned kernels, which keep the size of network representation space in terms of kernel numbers. The full space enables our method to accurately identify the lowest symbolic kernels in each iteration. 
We also test the effect of normalizing the kernels in each layer before clustering. Result in Fig.~\ref{fig:ablation_psm_normalize} suggests that normalization hurts the performance. Since our method relies on the interaction between kernels in each layer, normalizing kernels breaks the primitive relation, thus decreases the accuracy. 
\section{Feature Distillation of HRNet}
\begin{figure}[h!]
	\centering
	\includegraphics[scale=0.65]{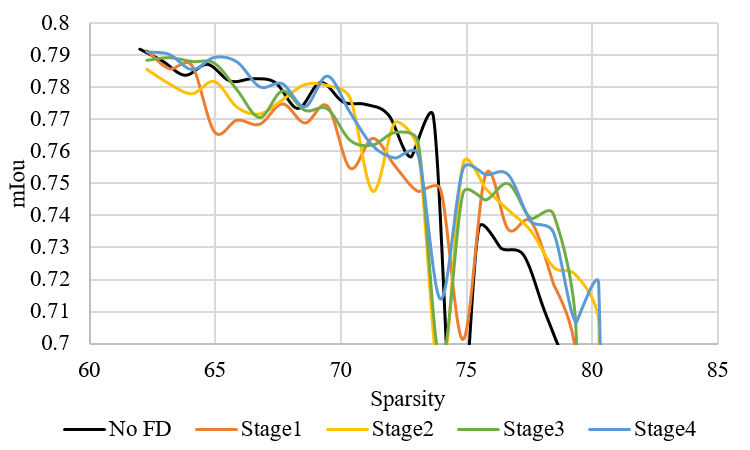}
	\caption{Variants of feature distillation for HRNet}
	\label{fig:hr_kd}
\end{figure}
As mentioned in Sec.~\ref{sec:FD}, features distillation of depth estimation cannot be equally applied on segmentation. Cost volume is a significant target for stereo depth estimation to transfer, while there is no such equivalent in segmentation. In low sparsity range, distilling the intermediate stage output results in worse accuracy. Fig.~\ref{fig:hr_kd} shows the distillation result of sparsity above 60\%. Larger stage denotes that the feature used for distillation is extracted from the layer closer to the final output. Stage1 feature preserves more characteristic of the input, while Stage4 feature is more like the final predicted class label. The results show that despite the accuracy of model using distillation can exceed model without distillation at some point, none of the distillation curves (Stage1 to Stage4 curves) constantly achieves better accuracy than none distillation model (No FD curve) below 75\% sparsity. At sparsity above 75\%, the accuracy drop becomes larger (yet still within 10\%) and unstable. In summary, feature distillation can still help to retain the accuracy of HRNet at high sparsity range, but the effect is much less significant than cost volume feature distillation (c.f Sec~\ref{sec:FD}) of stereo depth estimation networks. 
\section{Adversarial Criteria on Stereo Matching and Semantic Segmentation}
We also apply adversarial criteria on PSMNet and HRNet to validate our method. The results in Table~\ref{table:adv_psm} and~\ref{table:adv_hr} are coherent with Table~\ref{table:adv_2}. Adversarial criteria greatly affect the performance of KCP, which well demonstrates that the proposed original KCP is an effective method in terms of finding low representational kernels then remove them. 
\begin{table}[h!]
	\centering
	\caption{Results of adversarial criteria applied on PSMNet. $\mathcal{S}$ is set to 90. The results are extracted from different epoch during the training process. Non Adv. denotes original KCP. 10\% Adv. denotes that 10\% of the layers are randomly chosen to be pruned using adversarial criteria and vice versa.}
	\label{table:adv_psm}
	\begin{tabular}{ccccc}
		\toprule
		Spar.      & Non Adv. & 10\% Adv. & 20\% Adv. & 40\% Adv. \\ \midrule
		$\approx$10\% & 98.60\%  & 96.57\%   & 95.93\%   & 91.22\%   \\ \midrule
		$\approx$20\% & 98.51\%  & 94.46\%   & 90.39\%   & 75.26\%   \\ \midrule
		$\approx$30\% & 98.37\%  & 92.92\%   & 72.11\%   & 45.01\%   \\ \midrule
		$\approx$40\% & 98.30\%  & 79.10\%   & 35.47\%   & 13.87\%   \\ \midrule
		$\approx$50\% & 98.18\%  & 70.81\%   & 32.66\%   & 14.09\%   \\ \midrule
		$\approx$60\% & 97.77\%  & 37.17\%   & 14.10\%   &   -        \\ \midrule
		$\approx$70\% & 97.41\%  & 59.42\%   &    -       &    -       \\ \bottomrule
	\end{tabular}
\end{table}
\begin{table}[h!]
	\centering
	\caption{Results of adversarial criteria applied on HRNet. $\mathcal{S}$ is set to 30. The results are extracted from different epoch during the training process. Non Adv. denotes original KCP. 2\% Adv. denotes that 2\% of the layers are randomly chosen to be pruned using adversarial criteria and vice versa.}
	\label{table:adv_hr}
	\begin{tabular}{ccccc}
		\toprule
		Spar.       & Non Adv. & 2\% Adv. & 5\% Adv. & 10\% Adv. \\ \midrule
		$\approx$ 5\%  & 0.8111   & 0.7828   & 0.6589   & 0.6571    \\ \midrule
		$\approx$ 10\% & 0.8116   & 0.7517   & 0.6317   & 0.4938    \\ \midrule
		$\approx$ 15\% & 0.8109   & 0.7450    & 0.5866   & 0.2456    \\ \midrule
		$\approx$ 20\% & 0.8121   & 0.7125   & 0.2470    & 0.2028    \\ \midrule
		$\approx$ 30\% & 0.8096   & 0.6058   & -        & -         \\ \bottomrule
	\end{tabular}
\end{table}
\section{More Qualitative Results}
\begin{figure*}[h!]
	\centering
	\includegraphics[scale=0.2345]{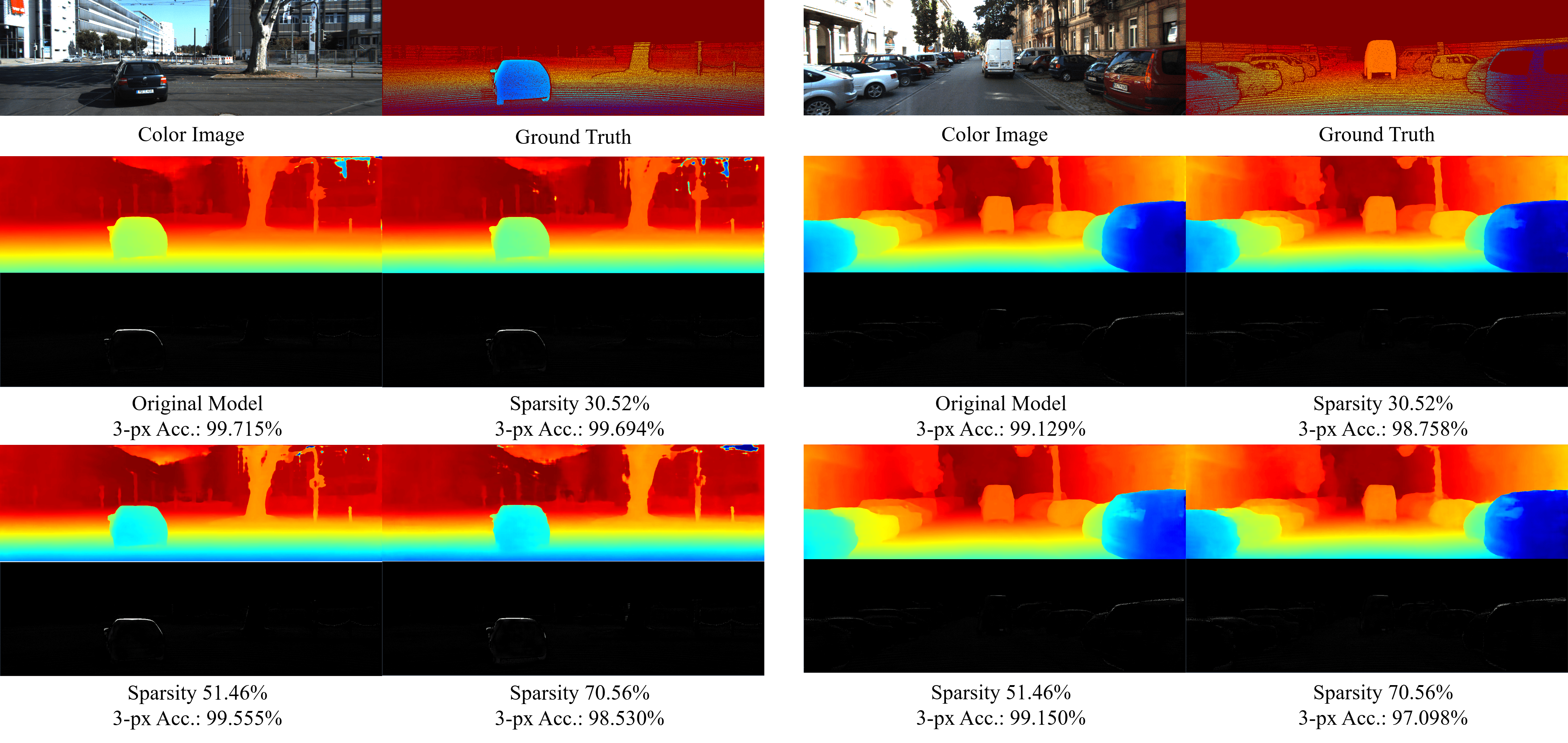}
	\caption{Results of our method of PSMNet on KITTI2015. We show the predicted depth map and the difference with ground truth (brightness indicates difference) from unpruned original model, 30\% sparsity, 50\% sparsity and 70\% sparsity. The 3-px accuracies are listed below each predicted results.}
	\label{fig:psmnet_vis2}
\end{figure*}

\begin{figure*}[h!]
	\centering
	\includegraphics[scale=0.2345]{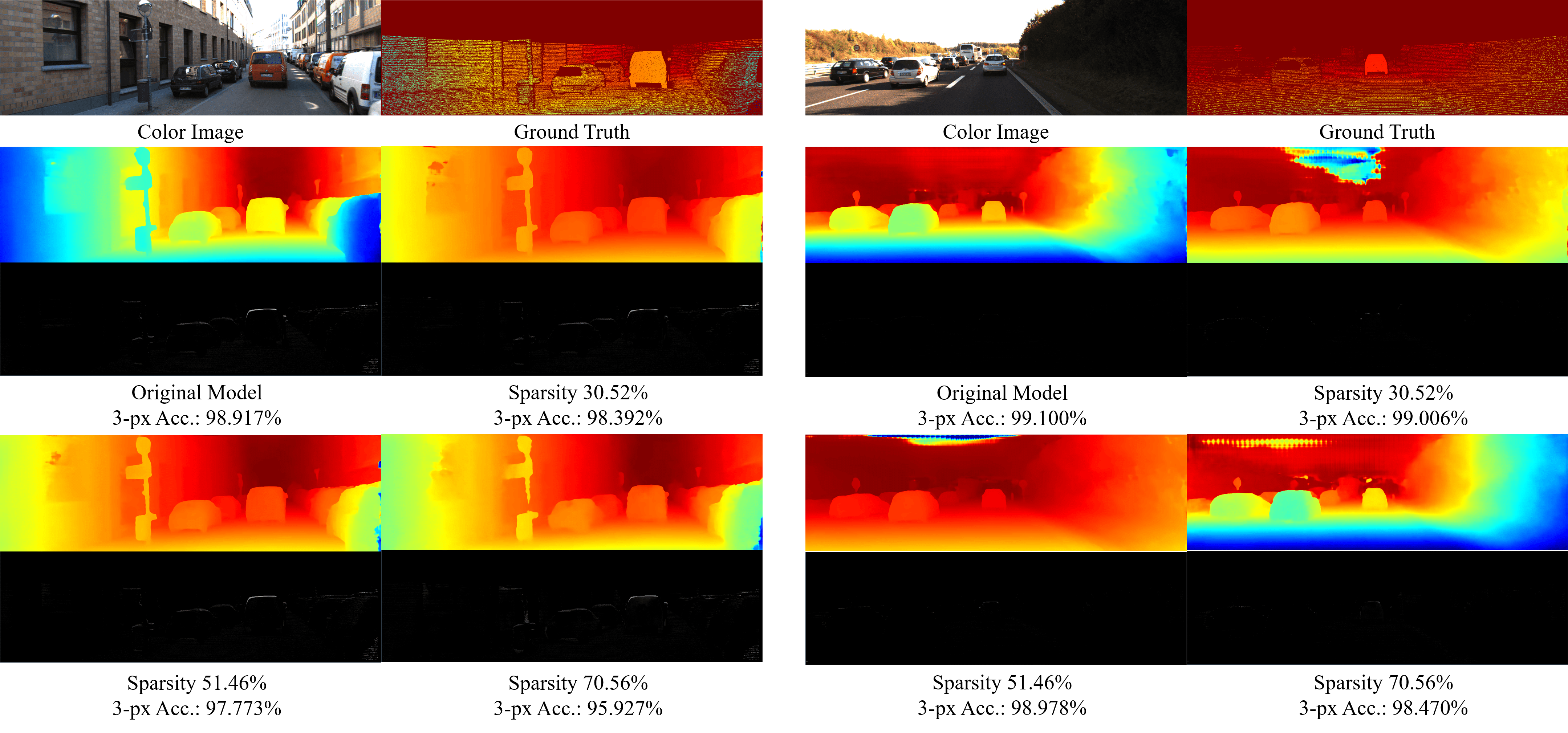}
	\caption{Results of our method of PSMNet on KITTI2015. We show the predicted depth map and the difference with ground truth (brightness indicates difference) from unpruned original model, 30\% sparsity, 50\% sparsity and 70\% sparsity. The 3-px accuracies are listed below each predicted results.}
	\label{fig:psmnet_vis4}
\end{figure*}

\begin{figure*}[h!]
	\centering
	\includegraphics[scale=0.2345]{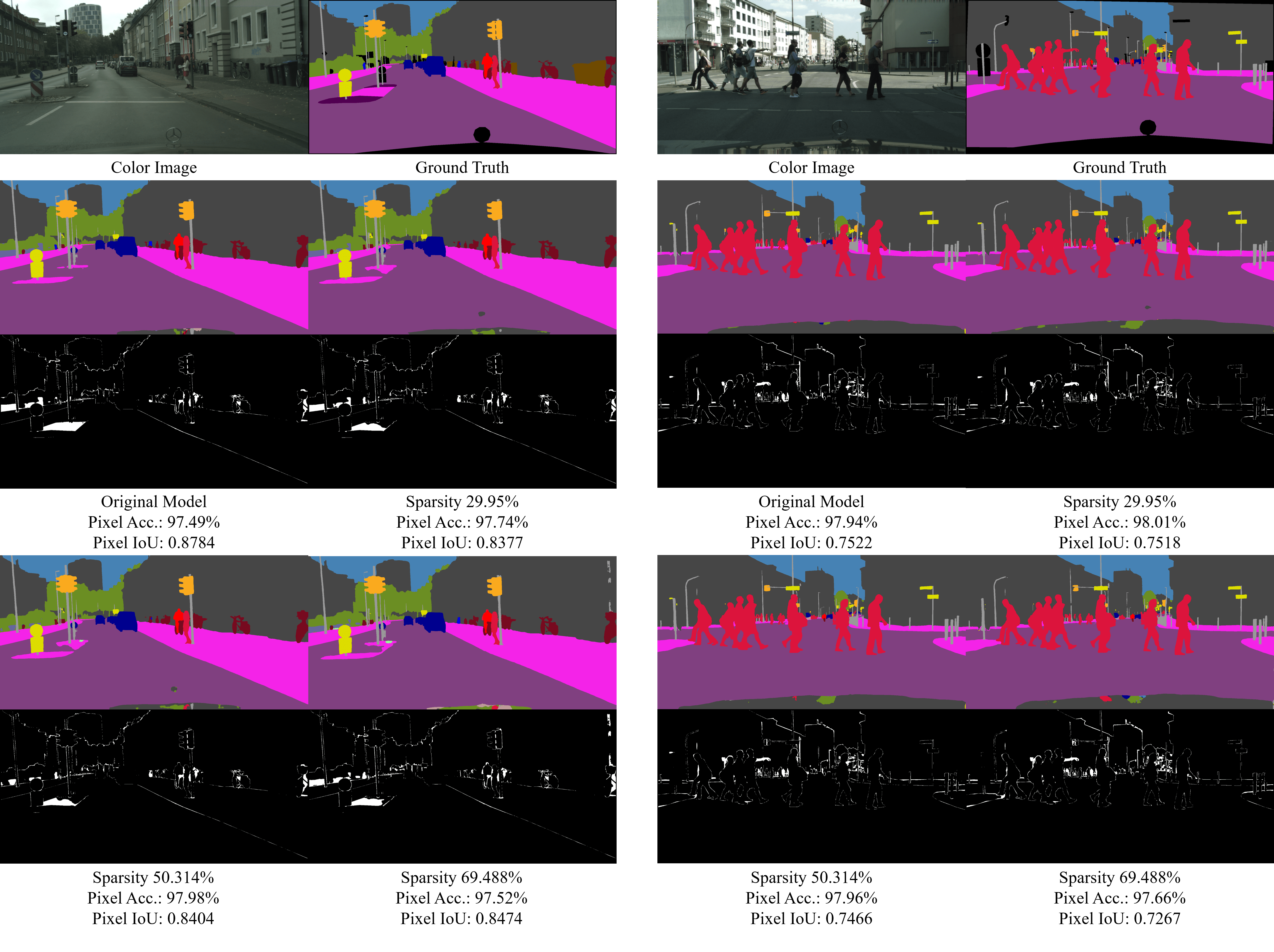}
	\caption{Results of our method of HRNet on Cityscapes. The predicted class label and the difference between ground truth are shown, (brightness indicates difference) including original model, 30\% sparsity, 50\% sparsity and 70\% sparsity.  The pixel accuracy and pixel IoU are listed below each figure.}
	\label{fig:hrnet_vis1}
\end{figure*}

\begin{figure*}[h!]
	\centering
	\includegraphics[scale=0.2345]{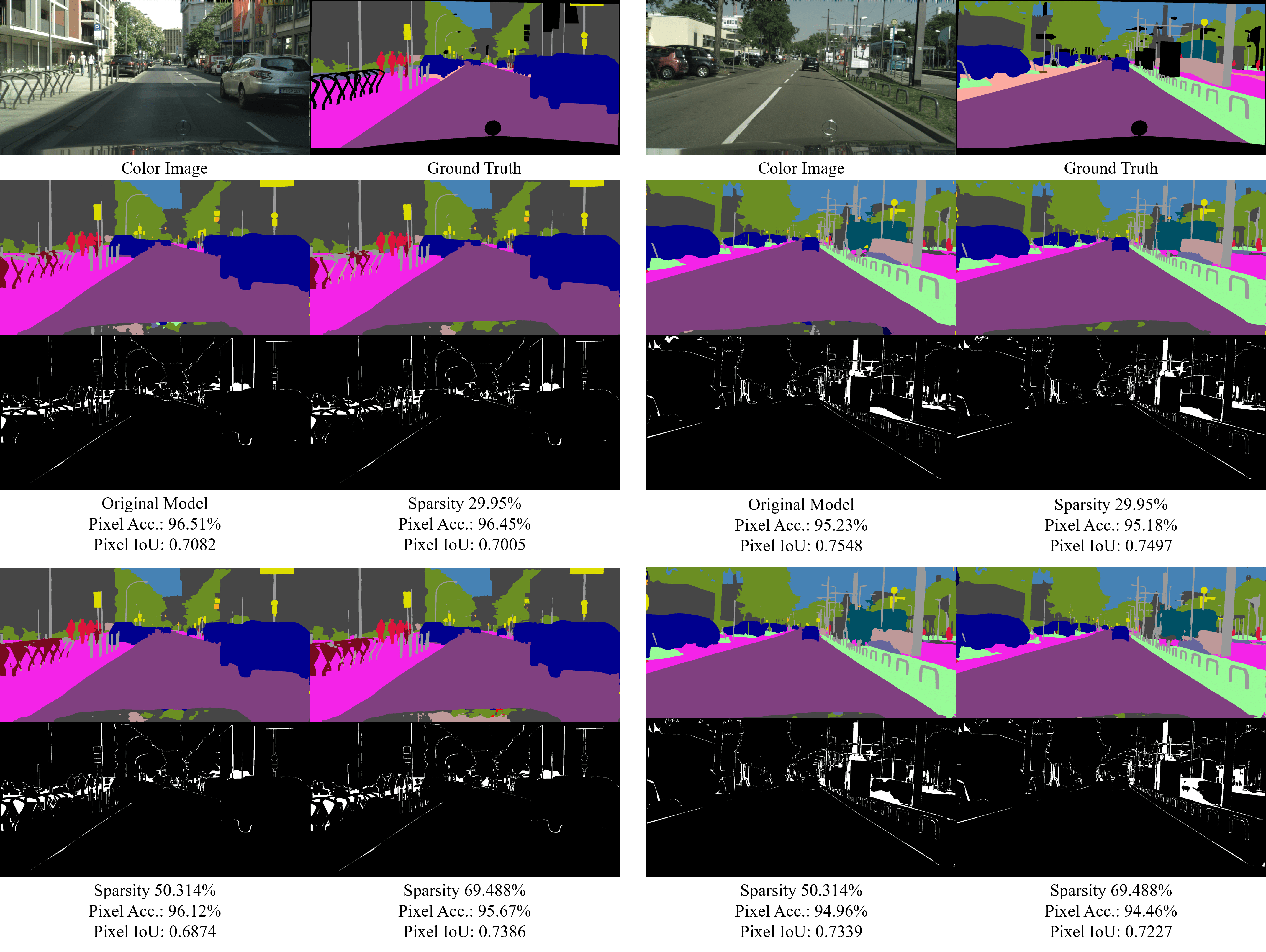}
	\caption{Results of our method of HRNet on Cityscapes. The predicted class label and the difference between ground truth are shown, (brightness indicates difference) including original model, 30\% sparsity, 50\% sparsity and 70\% sparsity.  The pixel accuracy and pixel IoU are listed below each figure.}
	\label{fig:hrnet_vis2}
\end{figure*}

\begin{figure*}[h!]
	\centering
	\includegraphics[scale=0.43]{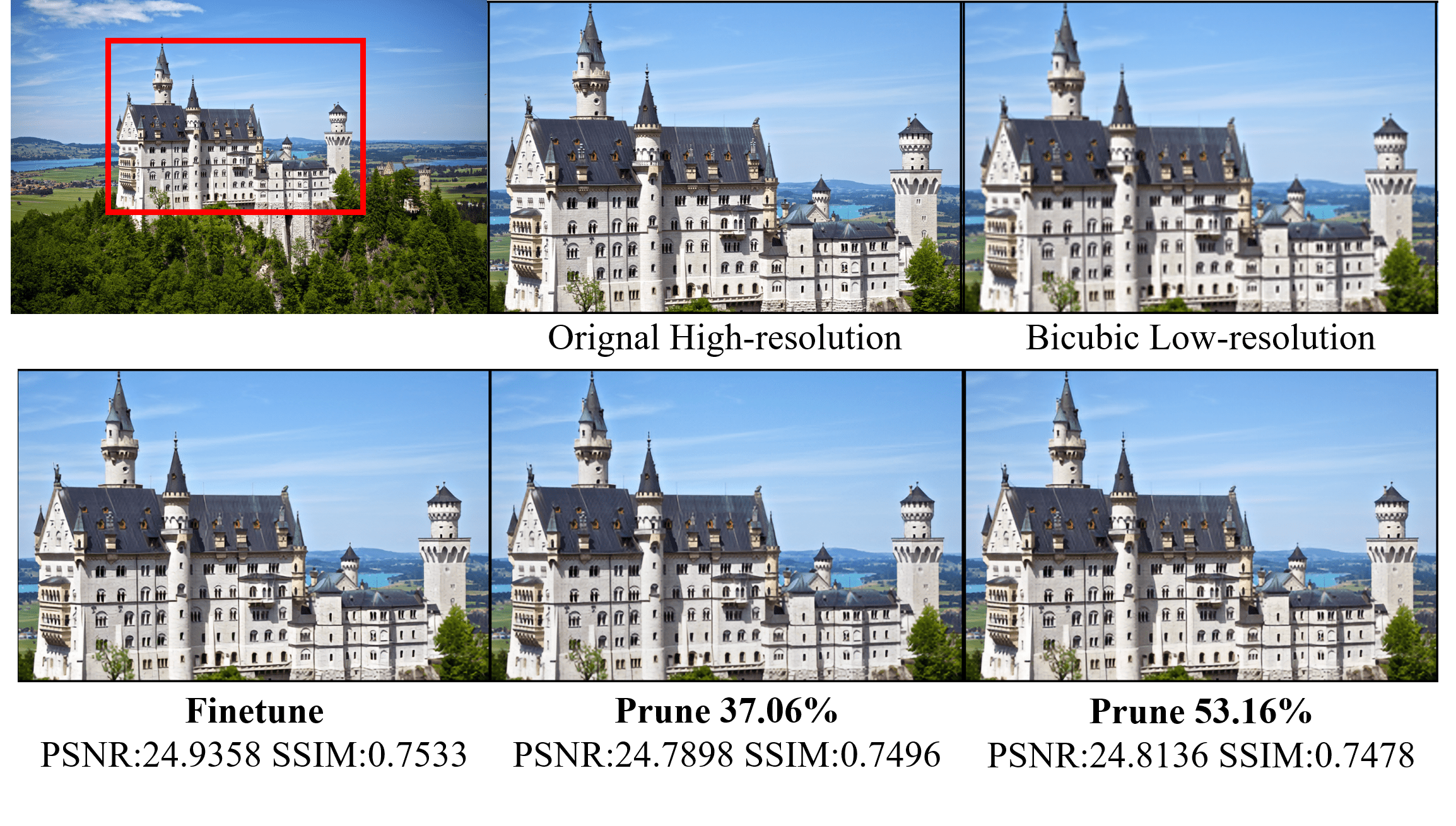}
	\caption{Corresponding high resolution image, bicubic-upsampled low resolution image and reconstruction results from different sparsity SRGAN generators. The image is from DIV2K validation set. Results are zoomed-in from the red block.}
	\label{fig:srgan2}
\end{figure*}

\begin{figure*}[h!]
	\centering
	\includegraphics[scale=0.43]{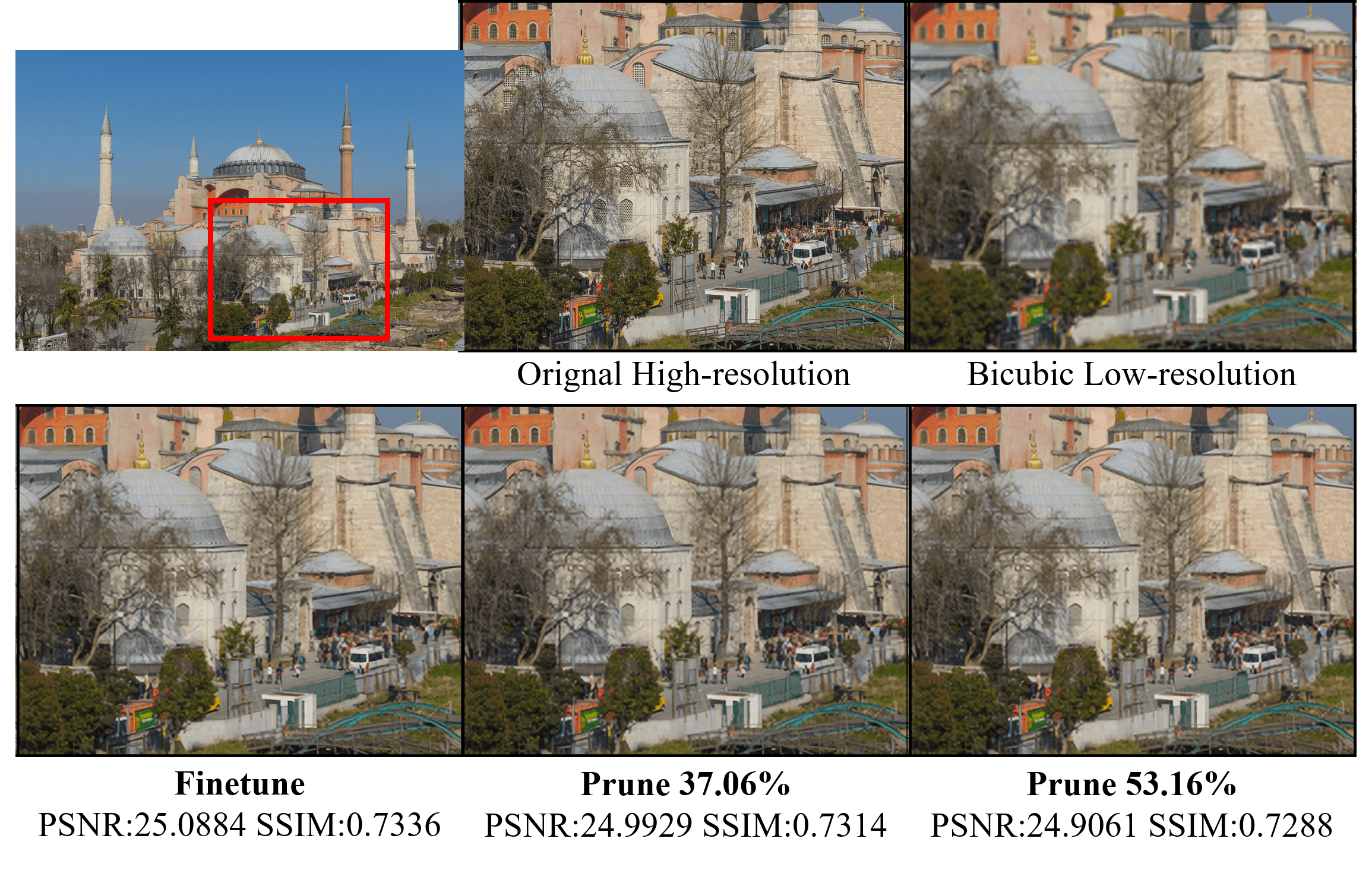}
	\caption{Corresponding high resolution image, bicubic-upsampled low resolution image and reconstruction results from different sparsity SRGAN generators. The image is from DIV2K validation set. Results are zoomed-in from the red block.}
	\label{fig:srgan3}
\end{figure*}

We provide more visualization results of KCP on different applications. Fig.~\ref{fig:psmnet_vis2} and~\ref{fig:psmnet_vis4} are predicted depth from KITTI2015 of PSMNet. The layout is identical with Fig~\ref{fig:psmnet_vis} in the main manuscript. Fig.~\ref{fig:hrnet_vis1} and~\ref{fig:hrnet_vis2} are predicted class labels from Cityscapes of HRNet. The layout is identical with Fig~\ref{fig:hrnet_vis} in the main manuscript. Fig.~\ref{fig:srgan2} and~\ref{fig:srgan3} are SRGAN reconstruction results of image from DIV2K dataset.  


\end{document}